# PREDICTING CUSTOMER CALL INTENT BY ANALYZING PHONE CALL TRANSCRIPTS BASED ON CNN FOR MULTI-CLASS CLASSIFICATION

Junmei Zhong and William Li

Marchex Inc
520 Pike Street, Seattle, WA, USA 98052

*ABSTRACT*

*Auto dealerships receive thousands of calls daily from customers interested in sales, service, vendors and jobseekers. With so many calls, it is very important for auto dealers tounderst and the intent of these calls to provide positive customer experiences that ensure customer satisfaction, deeper customer engagement to boost sales and revenue, and optimum allocation of agents or customer service representatives across the business. In this paper, we define the problem of customer phone call intent as a multi-class classification problem stemming from the large database of recorded phone call transcripts. To solve this problem, we develop a convolutional neural network (CNN)-based supervised learning model to classify the customer calls into four intent categories: sales, service, vendor or jobseeker. Experimental results show that with the thrust of our scalable data labeling method to provide sufficient training data, the CNN-based predictive model performs very well on long text classification according to tests that measure the model's quantitative metrics of F1-Score, precision, recall, and accuracy.*

## 1. INTRODUCTION

Auto dealership businesses grow by providing positive customer experiences. The more the customer scan actively engage, the better the enterprises grow. Every day, auto dealers receive many inbound phone calls, with the purposes of these calls typically ranging from sales inquiries, service requests, vendor questions, and job opportunities. Understanding call intent can help dealerships provide positive experiences to offer deeper customer engagement which could in turn boost sales and revenue, and optimum allocation of agents or customer service representatives to avoid understaffed or overstaffed situations. However, there are a lot of challenges in analyzing the massive datasets of phone calls. First, the dataset is too large to analyze manually and, second, different customers often use different words to express their intent in phone calls, challenging traditional NLP and machine learning algorithms. In this paper, we develop an artificial intelligence (AI)-based customer call intent prediction strategy to leverage the power of AI algorithms in semantically analyzing big transcripts of phone calls. Specifically, we train the convolutional neural networks (CNN)-based supervised learning model [1] with word embeddings to extract semantic features in the heterogeneous transcripts for classifying each phone call into one of the four intent categories.

CNN has been used successfully for many short text classification tasks such as sentiment analysis from Twitter [2], movie review analysis from sentence classification [3], and customer churn prediction by analyzing social media data, such as microblog data, for telecommunication industry [4]. Inspired by these widely adopted application examples of CNN, in this paper, we use CNN for long actual phone call transcript data analysis for call intent prediction. For this purpose, we collaborate with our client, one of the biggest auto dealerships in the United States, to use its actual customers' phone call data for building a model of call intent prediction that could produce real positive impact on the auto dealer's business. We use audio recordings of actual customer phone calls and apply our company's speech recognition techniques to translate the audio data into text data (transcripts) for analysis. Totraina CNN model with many parameters through supervised learning, a big challenge is how to annotate sufficient training examples in a cost-efficient way for building a reliable CNN predictive model. When we tested and analyzed existing CNN models [4] for our churn prediction work, we found out that the CNN models were not trained well enough due to too few training examples. As a result, the reported classification performance is not very satisfactory [4]. To solve the problem of insufficient training data for deep learning, our company has recently developed a scalable data labeling method to label the transcript data on the utterance level. With the power of this scalable data labeling tool, it only takes our four labelers approximately two weeks to gain sufficient training, validation, and testing examples.

Our call intent prediction system consists of four core components. First, we collect the phone call transcript data from the database. We only collect the caller channel transcripts since our domain knowledge of auto dealer enterprise customer service interactions informs us that customers' call intent signals are mainly contained in the caller channel, not in the agent channel. This has an added benefit since the use of the caller transcript can significantly reduce the amount of data for both data labeling and data analytics. Second, we label the ground-truth data as the training, validation, and testing examples, using our domain knowledge about call intent from the collected caller channel transcripts with the scalable data-labeling method. Third, we use natural language processing (NLP) algorithms for document tokenization and words' embed dedvector representation. The fourth and final component is training the multi-class text classification model for call intent prediction using the CNN algorithm. Experimental results show that when sufficient training examples are provided, our CNN model generates state-of-the art performance in text classification for customer call intent prediction according to the quantitative metrics of accuracy, F1-score, precision and recall. Our contributions are demonstrated in the following two aspects:

- We conduct the analysis on actual customers' phone calls for customer call intent prediction. To the best of our knowledge, it is the first proposal to use the caller channel transcripts from the caller-agent two channel phone conversations for this prediction task. This not only reduces the amount of data for both fast data labeling and data analysis, but also improves the prediction performance.
- We develop AI algorithms including NLP and CNN algorithms for multi-class call intent prediction from transcripts of customer phone calls with sufficient training data to make the CNN model optimal.

The rest of the paper is organized as follows. In Section 2, we discuss the research methodology in detail. In Section 3, we present the experimental results, and we conclude the paper with some discussions and the direction of future work in Section 4.

## 2. RESEARCH METHODOLOGY

There are three components involved in this research methodology for customer call intent prediction for auto dealerships: Problem definition and text data preparation from phone calls, embedded vector representation for tokens/words in each transcript, and the CNN-based supervised learning algorithm for multi-class call intent prediction from the classification of transcripts.

### 2.1 Problem Definition and Text Data Preparation

When we want to apply any AI algorithm to practical problems in order to generate real impact on reducing costs or growing business, it is very important to first define the underlying business problem according to the specific domain knowledge, and then figure out what kind of data should be collected and prepared from the databaseso that AI algorithms can be used to leverage the insights from the dataset and generate actionable intelligence by training reliable AI models.

Our phone call data is from an auto dealer's daily customer service phone calls. Each phone call consists of information from two channel communications: the agent channel and the caller channel. Each agent or caller channel consists of a sequence of utterances (usually sentences in the transcripts). According to our domain knowledge of auto dealership business, the customer's call intent information is mostly contained in the caller channel, rarely is it found in the agent channel. Knowing the characteristics of our phone call data and our client's business goal of understanding the customers' main intent in each call, we define the customer call intent prediction problem as a multi-class text classification problem: hiring, sales, service, and vendor. Furthermore, for this multi-class classification problem, according to the client's business request, each call is exclusively classified into one of the 4 classes without having multiple intents. This is due to the specific business scenarios. In people's daily chat over the phone, it is highly possible for people to talk about multiple things, but for auto dealers' customer calls, it is seldom for job seekers to ask vendor questions, inquire auto sales or request auto service. In the same way, calls inquiring for auto sales seldom seek jobs in the dealer, ask vendor questions, or request auto services. For this classification purpose, we propose to only collect data from the caller channel transcripts for all the phone calls. This not only reduces the amount of data to review when labeling the data, but also has the potential to improve the prediction performance at model's training and prediction stages by discarding the irrelevant agent channel. We label the caller channel transcripts and use them as the training data.

### 2.2 Embedded Vector Representation for Documents

The transcript of each call's caller channel in the training examples is taken to be a document. When using machine learning algorithms for text classification, documents need to be first tokenized into individual words or tokens. For traditional machine learning,documents can be represented as feature vectors of tokens using the TF-IDF weighting method or binary method, but it has been concluded that this way is not efficient enough. In deep learning, tokens are represented as embedded vectors for extracting semantic features for text classification, text understanding, or text generation. So, we first tokenize each document into a collection of terms. After some pre-processing such as cleaning, removal of stop words, and normalization for the tokens, is conducted, we then represent each document as a matrix of pretrained word embeddings of word2Vec [5] or GloVe [6].

**2.2.1 The Bag of Words (BOW) Method**

The BOW method makes use of tokenized individual words and/or N-Grams (N>=1)in a corpus as features for a document's vector representation, which is usually called a feature vector in machine learning and pattern recognition. If N is equal to 1, the N-Grams is called unigram. Usually at most we use bi-Gramsor triple-Grams for practical considerations. For individual tokens, the BOW method usually has both binary representation and TF-IDF weighting representation to get the feature values. The binary representation only considers the presence and absence of individual tokens without taking the frequency of their occurrences into account. On the other hand, the TF-IDF weighting method takes the product of two statistics: the term frequency and inverse document frequency. The term frequency is simply the number of occurrences of the term $t$ appearing in a document $d$. This is based on the idea that the more frequent a token appears in the document, the more important the token is in representing the topics of the document, and it is usually calculated in the following augmented way:

$$tf(t, d) = 0.5 + 0.5 * \frac{f_{t,d}}{\max\{f_{t',d} : t' \in d\}} \qquad (1)$$

where $f_{t,d}$ denotes the frequency of term $t$ in document $d$. At the same time, the inverse document frequency is calculated in the following way:

$$idf(t, D) = \log\left(\frac{N}{|\{d \in D : t \in d\}| + 1}\right) \qquad (2)$$

With

- N the number of documents in the corpus D, N= |D|
- The denominator $|\{d \in D : t \in d\}| + 1$ is the number of documents where the term $t$ appears.

The inverse document frequency is used to offset the impact of common words without having specialty. But the BOW method usually has some limitations. The first limitation is that it does not consider the order information of terms for document classification, but only considers the occurrences of individual terms, which is not true from both semantic and syntactic points of view. As a result, documents with different semantic meanings is easy to be classified into the same class only if they contain the same terms or vice versa. The other limitation is the high dimensionality. Corpora generally have at least thousands of words. In addition to this, if the 2-grams and 3-grams are included, the number of features for document increases significantly. It could generate an even more sparse term-document matrix leading to significantly increased demand of training data and potential over fitting problem for supervised learning.

**2.2.2 Word2Vec for Word Embeddings**

The BOW-based vector representation is obviously not an efficient method to capture the semantic information from documents with additional limitations of high dimensionality and sparsity, so researchers have proposed different methods to represent documents and words in an embedded low-dimensional continuous vector space. The word2vec algorithm developed at Google is such a distributed representation learning algorithm to extract both semantic and syntactic information for individual words in a sentence. It consists of a few related models that are used to produce the distributed representation of word embeddings. These models are the continuous bag-of-words (CBOW) and the skip-gram as shown in Figure 1. The CBOW model predicts the current word from its surrounding context words within a window centered at the current word, on the other hand, while the skip-gram model predicts the surrounding context words within a window for this current word. The word2vec model is an unsupervised learning algorithm which can be trained with the hierarchical softmax and/or negative sampling method to reduce the computational complexity and make the learning process practical. These two models are the two-layer shallow neural networks. Word2vec takes as its inputs the high dimensional one-hot vectors of words in the corpus and produces a vector space of several hundred dimensions which are much smaller than the size of the vocabulary, such that each unique word in the corpus is represented by a continuous dense vector in the embedded vector space. A very salient feature of this kind of vector representation with the word embeddings is that word vectors are close to each other for semantically similar words, and they can be inferred from each other. This offers great benefits for semantic document analysis with traditional machine learning and deep learning algorithms.

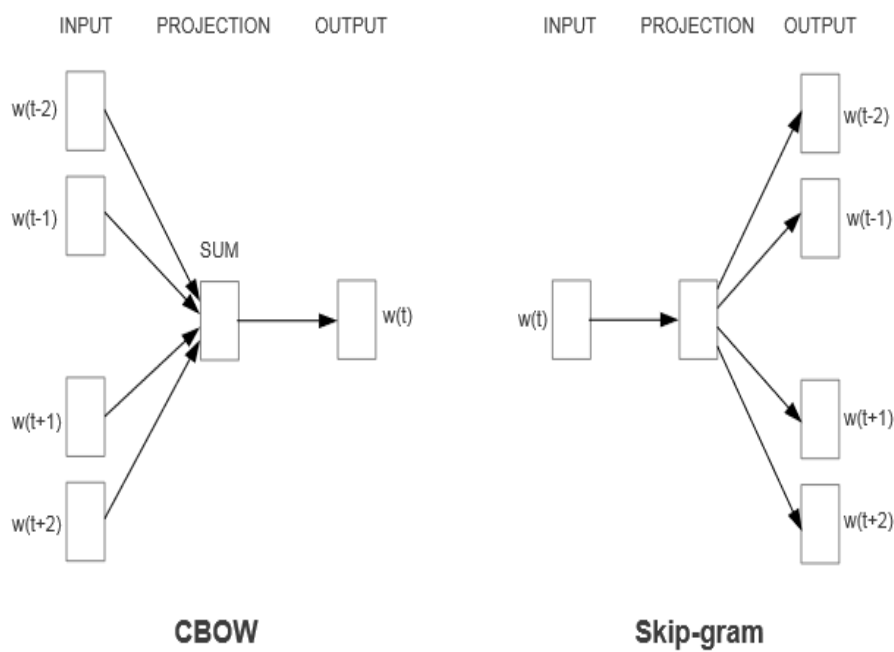

Figure 1. The illustration of CBOW and Skip-gram models in Word2Vec with courtesy of Mikolov etc. [5]

**2.2.3 GloVe for Word Embeddings**

GloVe is another kind of global vectors as word embeddings for word representation [6]. It is also a kind of unsupervised learning to learn word embeddings by collecting the word-word co-occurrence statistics from a corpus. The word-word co-occurrence statistics is represented in the form of words' co-occurrence matrix such that each element of the matrix represents how often or the probability a word $w_i$ appears in the context of another word $w_j$, and the context is often specified by a window. An example from the original GloVe paper [6] is given here: Consider the co-occurrence probabilities for given words "*ice*" and "*steam*" and various probe words "*solid*", "*gas*", "*water*", and "*fashion*" from the vocabulary. The actual probabilities of the co-occurrences are from a corpus with 6 billion words. The word "*ice*" co-occurs more frequently with "*solid*" than with "*gas*", whereas "*steam*" co-occurs more frequently with "*gas*" than with "*solid*". Both words co-occur with their shared property "*water*" frequently, and both co-occur with the unrelated word "*fashion*" infrequently. The ratio of probabilities encodes some crude form of semantic meaning.

The motivation of training the GloVe is to learn embedded word vectors from corpus such that for any two words, $w_i$ and $w_j$, the dot product of their corresponding embedded vectors $v_i$ and $v_j$ is equal to the logarithm of their probability of co-occurrence:

$$w_i^T w_j = \log(X_{i,j}) - \log(X_i) \tag{3}$$

This relationship can be further refined to satisfy the symmetric property of distances by introducing two bias items to replace the evidence term $\log(X_i)$:

$$w_i^T w_j + b_i + b_j = \log(X_{i,j}) \tag{4}$$

By minimizing the least-squares cost function J:

$$J = \sum_{i=1}^{V} \sum_{j=1}^{V} f(x_{ij})(w_i^T w_j + b_i + b_j - \log X_{ij})^2 \tag{5}$$

with $f(.)$ being a weighting function to help prevent learning only from extremely common word pairs and V denoting the vocabulary size of the corpus, we can learn the embedded vectors for all words in the corpus, generating a word vector space with meaningful substructure.

| Probability and Ratio | k = solid | k = gas | k = water | k = fashion |
|---|---|---|---|---|
| $P(k\|ice)$ | $1.9 \times 10^{-4}$ | $6.6 \times 10^{-5}$ | $3.0 \times 10^{-3}$ | $1.7 \times 10^{-5}$ |
| $P(k\|steam)$ | $2.2 \times 10^{-5}$ | $7.8 \times 10^{-4}$ | $2.2 \times 10^{-3}$ | $1.8 \times 10^{-5}$ |
| $P(k\|ice)/P(k\|steam)$ | 8.9 | $8.5 \times 10^{-2}$ | 1.36 | 0.96 |

Figure 2. An illustration of the word-word co-occurrence matrix [6]

The resulting embeddings possess interesting linear substructures of the words in the embedded vector space from the vector distance. As shown in the left panel of Figure 3, the vector distances for all man-woman pairs are similar. For example, the vector distances among the word pairs of (aunt, uncle), (woman, man), (madam, sir), and (heiress, heir), look like "vertical" substructures and are similar to each other. Also, from the right panel of Figure 3, it is easy to see the similar situation among the pairs of company element and CEO element with "horizontal" substructures. Like word2Vec embeddings, this kind of representation is very useful for many machine learning tasks with deep learning.

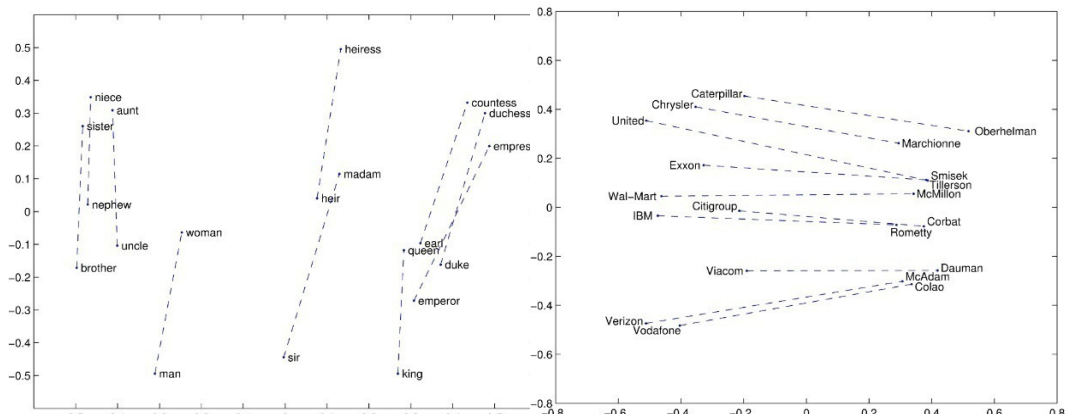

Figure 3. Substructures of Man-Woman (left) and Company-CEO (right) [6]

## 2.3. Training the CNN Model for Text Classification

CNN is one of the deep learning algorithms and it integrates the automatic feature extraction, feature selection and pattern classification in a single architecture. The automatic feature extraction is accomplished through the convolutional process with different sizes of filters to get the feature maps through convolutions while the feature selection is accomplished through the max-pooling on the feature maps. The pattern classification task is done through the fully connected layer in the CNN architecture. As depicted by Figure 4, for text classification, the CNN usually takes as inputs the matrix of the word embeddings of the sentence by stacking the words' vectors according to the order of the words in the sentence. If different word embeddings are used, there will have multiple channels for the inputs of CNN. Typically, the embeddings can be word2Vec and Glove. Each channel of the texts can be represented as a matrix, in which, the rows represent the sequence of tokens or words in a sentence, and each row is a word's embedding vector and the number of columns of the matrix is the dimension size of the embeddings. The matrix is convolved with some filters of different sizes such as 3, 4, 5, but with the same dimension as the words' embeddings.

The principal idea of CNN for text classification with different sizes of filters is to extract the N-Grams features by the sizes of filters. Let's assume the filter size is $m$, sentence length is $l$, dimensionality of word embeddings is $d$, then the sentence matrix $x \in R^{l \times d}$ and the filter can be represented as a matrix $w \in R^{m \times d}$. During the convolution process, each of the filters gradually moves along the sequence of words. At each position, the filter covers a portion of the words' vector matrix, i.e., m words' vectors in the matrix, and the point wise multiplication of the filter with the covered vector matrix is taken, and the multiplication results are summed up. This sum is then taken by an activation function such as the rectified linear unit (Relu) together with a bias term $b \in R$ to generate a feature value mathematically represented in the following formula:

$$c_i = f(w \cdot x_{i:i+m-1} + b) \tag{6}$$

where the dot operation between matrix w and x is the element-wise multiplication operation. After the convolution process is done for one filter, a list of feature values is obtained as the feature map, $c = [c_1, c_2, c_3, \ldots, c_{l-m+1}]$. Then, the max-pooling operation continues to take the maximum value from the feature map as the feature value of the filter's convolutional result with the sentence. When all filters are applied for convolution with the sentence's vector matrix, a list of feature values is obtained as the feature vector representation for the input sentence data. This feature extraction and selection process with the convolution and max-pooling operations makes the length of the final feature vector independent of the input sentence length and the filter sizes. The length of the final feature vector is only dependent on the number of filters used for convolution. The final step in the CNN architecture is a full connection layer with dropout and regularization from the final feature vector to the output layer. The classification result of a sample is obtained by the softmax function applied to the output layer for multi-class classification. The number of neurons in the output layer depends on the number of classes in the classification task. For our 4-class classification problem, there are 4 neurons in the output layer.

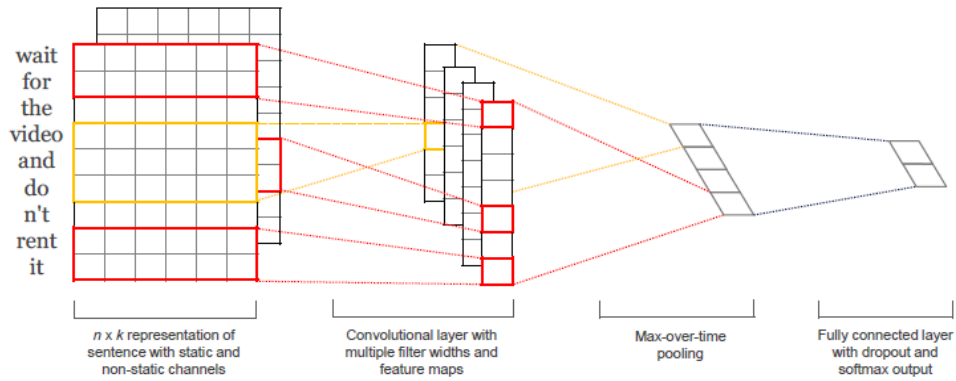

Figure 4. The CNN architecture for text classification with courtesy of Yoon Kim [1].

## 3. EXPERIMENTAL RESULTS AND ANALYSIS

To train are liable CNN-based multi-class classification model for customer call intent prediction, we annotate about 2,200 samples of transcripts from the caller channels of phone call data for each of the 4 intent classes, and these annotated examples are used as the training and validation. We also label 150 testing examples for each intent class to test the trained model.

When CNN is used for text classification, its typical application is for short text or sentence classification such as for movie review classification, service or product's customer review (CV) classification, and so on. But in our CNN-based call intent prediction model, the text data for the CNN is the whole caller channel's transcript of each phone call. We extend the input of CNN from short sentence to long document for text classification by taking the whole caller channel transcript as a "long sentence" rather than doing the convolution for the individual sentences in the transcript. So, we get one single feature vector for each transcript from the convolution. The inputs of the words' vectors for the CNN algorithm can be either the pre-trained word embeddings of word2Vec or GloVe. For feature extraction through convolutions, we use 3 different convolving filter sizes, 3, 4, and 5, and we use 128 filters for each filter size for feature extraction in the convolutional process. For the hyper parameters of CNN, our final settings are: batch size 32, epoch size 50, the dimension of word embeddings 100, dropout 0.5, decay coefficient 2.5, and $l_2$-norm-lambda 0.05 is used for the cross-entropy loss regularization coefficient. It is also found out that the optimizer RMSProp works best among AdamOptimizer, AdagradOptimizer, AdagradDAOOptimizer, and AdadeltaOptimizer. The CNN algorithm is implemented in Python with the tools of Tensor flow, Sk-learn, and Numpy. In this paper, we use the quantitative metrics accuracy, precision, recall, and F1-score to measure the performance of the model on the testing dataset, and they are calculated in the following way:

$$Precision = \frac{tp}{tp+fp} * 100 \qquad (7)$$

$$Recall = \frac{tp}{tp+fn} * 100 \qquad (8)$$

$$F_{1\_score} = 2 \cdot \frac{precision*recall}{precision+recall} * 100 \qquad (9)$$

$$Accuracy = \frac{tp+tn}{tp+fp+tn+fn} * 100 \qquad (10)$$

where *tp* denotes true positives, *fp* denotes false positives, *tn* denotes true negatives, and *fn* denotes false negatives. These metrics are multiplied by 100 in this paper for clarity consideration.

From the previous work [1, 4], it has been verified that the CNN model with the pre-trained word embeddings generally outperforms the traditional machine learning models such as the naïve Bayes and SVM, for which the bag of word (BOW) based vector space representation is used for document representation. So, in this work, we do not train those traditional machine learning models again. Table 1 lists the previous prediction results of the CNN model with 4 different word embeddings for churn prediction using the microblog social media data [4]. It is clear to see that the difference of prediction performance between using the word embeddings of word2Vec and GloVe. So, in this paper, we only use the GloVe as the word embeddings to train our CNN model for customer call intent prediction from phone call transcripts. Table 2 lists the previous prediction results of CNN + different logic rules for distilling the knowledge into the neural networks [4]. It demonstrates that when the logic rules are added [7], better prediction performance can be achieved. Table 3 lists our predictive model's averaged prediction performance on about 150 testing examples for each of the 4 intent classes for phone call data analysis. Table 4 lists the performance report in terms of accuracy for 6-class classification using different deep learning models[1] on the TREC dataset with 10 words in each sentence. Table 5 lists our model's prediction results for each of the 4 intent classes. Table 6 displays the confusion matrix of our model on the 4 intent classes. By comparing Table 2 with Table 3, it is easy to see that our CNN-based predictive model outperforms the published models and gain about 10for both precision and F1-score and gain 8 for recall. Even though we do not use additional logic rules [7] to distill the knowledge into the neural networks, we still get much better prediction performance. Our analysis reveals that the outstanding prediction performance is mainly attributed to the fact that we can collect much more training examples to train the CNN model than that for the previous work [4]. Those models in Table 2 use much less training examples for training those CNN models. As a result, for the curse of dimensionality, it is hard to say those CNN-based models in Table 2 are optimized well enough. This experiment confirms the common sense that for deep learning, how to get sufficient high-quality training data is crucial to the success! It also points out the main effort in training deep learning models is to first collect sufficient training data, not struggling on the selection of the best deep learning algorithm. When comparing Table 3 with Table 4, we can see that our CNN model for long text classification performs comparatively with the other outstanding deep learning models for short sentence classification, generating state-of-the-art multi-class classification performance for long texts. From Tables 5&6, we can see that our CNN model works very well on predicting intents of job seeking and vendor.

Table 1. The reported churn prediction results of CNN for microblog data analysis using different word embeddings [4].

| Input Vector(s) | F1-Score |
|---|---|
| Random embeddings | 77.13 |
| CBOW | 79.89 |
| Skip-Gram | 79.55 |
| GloVe | 80.67 |

Table 2. The reported churn prediction results of CNN + three logic rules for microblog data analysis with/without using word embeddings [4]

| Models | F1-Score | Precision | Recall |
|---|---|---|---|
| CNN | 77.13 | 75.36 | 79.00 |
| CNN+pretrained | 80.67 | 79.28 | 82.11 |
| CNN+pretrained+ "but" rule | 81.95 | 80.84 | 83.09 |
| CNN+pretrained+ "switch from" rule | 80.92 | 79.74 | 82.14 |
| CNN+pretrained+ "switch to" rule | 82.60 | 80.89 | 84.39 |
| CNN+pretrained+ All the 3 rules | 83.85 | 82.56 | 85.18 |

Table 3. The average prediction results of our CNN model for the 4 intent classes

| Model | F1-Score | Precision | Recall | Accuracy |
|---|---|---|---|---|
| CNN + GloVe | 93 | 93 | 93 | 92.7 |

Table 4. The reported 6-class classification accuracy of CNN [1] and DCNN [8] on TREC dataset

| model | CNN-static | CNN-non-static | CNN-multi-channel | DCNN |
|---|---|---|---|---|
| accuracy | 93 | 93.6 | 92.2 | 93.0 |

Table 5. The prediction results of our CNN model for each of the 4 classes

| Class | F1-Score | Precision | Recall | Support |
|---|---|---|---|---|
| hiring | 98 | 100 | 95 | 151 |
| sales | 88 | 89 | 87 | 151 |
| service | 87 | 85 | 89 | 150 |
| vendor | 98 | 97 | 99 | 150 |

Computer Science & Information Technology (CS & IT) 19

Table 6. The confusion matrix of our CNN model for the 4 intent classes

| | | | | |
|---|---|---|---|---|
| hiring | 144 | 0 | 4 | 3 |
| sales | 0 | 131 | 19 | 1 |
| service | 0 | 15 | 134 | 1 |
| vendor | 0 | 1 | 0 | 149 |

## 4. CONCLUSION AND FUTURE WORK

In this paper, we develop a CNN-based predictive model for auto dealership customer call intent prediction. Experimental results demonstrate that the CNN algorithm with sufficient training data generates state-of-the-art prediction performance. In the future, we are going to investigate other deep learning algorithms in intent prediction.

**ACKNOWLEDGEMENTS**

The authors would like to thank the proofreading of Brian Craig and Jana Baker, which greatly improves the quality of this paper.

**AUTHORS**

**Junmei Zhong** received the Ph.D. degree from Electrical & Electronic Engineering, The University of Hong Kong in 2000, where he received the prize of "Certificate of Merits for Excellent Paper", awarded by IEEE Hong Kong Section and Motorola Inc, Dec. 1998, the Master's degree in Computer Science, Nankai University, Tianjin, China, in 1993, where he received the "Award of Excellent Thesis",the B.Sc. degree in Computer Science from Dalian University of Technology, China, in 1988. He is now the Chief AI Scientist at Marchex Inc. His R&D interests include Applied mathematics, machine learning, data mining, deep learning for NLP and computer vision, text mining, graph theory, knowledge graph, digital advertising, medical CT& MR imaging,medical health record analysis, signal processing, wavelets, image recognition, and pattern recognition with more than 20 publications on prestigious journals and peer-reviewed conference proceedings. Dr. Zhong was the research faculty in University of Rochester, NY and Assistant Professor in Cincinnati Children's Hospital Medical Center from 2002 to 2006. 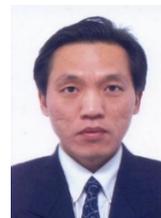

**William Li** is a Senior technology executive, currently the Vice President of Engineering at Marchex Inc, leading conversational intelligence strategy and product development. Before joined Marchex, he served as Director and Architect in DoubleClick, Microsoft and other companies managing multiple engineering groups. He founded and Led R&D divisions, AI Labs and Innovation Centers to drive artificial intelligence innovations. His recent work also includes Big Data mining, Computer Visions, Natural Language Processing research with a wide range of business applications. He is also a current member of Forbes Technology Council and founding member of IEEE Computer Society STC on Autonomous Driving. 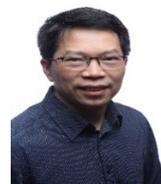